\newif\ifarxiv
\arxivtrue  

\ifarxiv
    \documentclass[11pt]{article}
    \usepackage{authblk}
    \usepackage[round]{natbib}
    \let\cite\citealp
\else
    \documentclass[final,3p,times,twocolumn]{elsarticle}
\fi
  
\usepackage{amssymb}       
\usepackage{amsmath}       
\usepackage{amsthm}        
\usepackage{amsfonts}
\usepackage{hyperref}      
\usepackage{xcolor}        

\usepackage{pdflscape}     
\usepackage{graphicx}      
\usepackage{subcaption}

\usepackage{multirow}      %
\usepackage{adjustbox}     
\usepackage{booktabs}      
\usepackage{siunitx}       
\usepackage{rotating}      
\usepackage{enumitem}      

\usepackage{algorithm}
\usepackage{algorithmic}   
\usepackage{verbatim}      

\usepackage{tabularx}      
\usepackage{array}

\newcommand{\OUTPUT}{\item[\textbf{Output:}]}
\newtheorem{theorem}{Theorem}
\newtheorem{cor}{Corollary}

\newtheorem{obs}{Observation}

\hypersetup{
            colorlinks = true,               
            urlcolor   = defaultURLColour,   
            citecolor  = defaultCiteLColour, 
           }

\definecolor{defaultURLColour}  {RGB}{127, 0, 255}
\definecolor{defaultCiteLColour}{RGB}{127, 0, 255}
\definecolor{darkgray}          {RGB}{128, 128, 128}  
\definecolor{midgray}           {RGB}{192, 192, 192}  
\definecolor{lightgray}         {RGB}{224, 224, 224}  

\date{} 					

\begin{document}

\ifarxiv
\title{Shapley-Inspired Feature Weighting in $k$-means with No Additional Hyperparameters}
\date{}
\author{Richard J. Fawley\thanks{\texttt{rf23433@essex.ac.uk}} \and Renato Cordeiro de Amorim\thanks{Corresponding author, \texttt{r.amorim@essex.ac.uk}}}
\affil{\small School of Computer Science and Electronic Engineering, University of Essex, Wivenhoe, UK.}
\maketitle
\begin{abstract}
Clustering algorithms often assume all features contribute equally to the data structure, an assumption that usually fails in high-dimensional or noisy settings. Feature weighting methods can address this, but most require additional parameter tuning. We propose SHARK (Shapley Reweighted $k$-means), a feature-weighted clustering algorithm motivated by the use of Shapley values from cooperative game theory to quantify feature relevance, which requires no additional parameters beyond those in $k$-means. We prove that the $k$-means objective can be decomposed into a sum of per-feature Shapley values, providing an axiomatic foundation for unsupervised feature relevance and reducing Shapley computation from exponential to polynomial time. SHARK iteratively re-weights features by the inverse of their Shapley contribution, emphasising informative dimensions and down-weighting irrelevant ones, and is equivalent to replacing the arithmetic mean of 
feature dispersions with their harmonic mean. Experiments on synthetic and real-world data sets show that SHARK consistently matches or outperforms existing methods, achieving superior robustness and accuracy, particularly in scenarios where noise may be present.
\noindent\textbf{Software:} https://github.com/rickfawley/SHARK.
    \end{abstract}

\fi



\section{Introduction}
\label{sec:intro}

Clustering algorithms aim to form groups of data points, known as clusters, such that points within the same cluster are similar, while points between clusters are dissimilar, according to some pre-defined dissimilarity measure. They have been applied successfully in diverse domains, including genomics, natural language processing, image analysis, network intrusion detection, and general data mining (\cite{ran2023comprehensive,mahdi2021scalable,ikotun2023k}).

A key advantage of clustering algorithms is that they follow the unsupervised learning paradigm: they operate without labeled samples, relying entirely on the data. This is especially valuable when labeling is costly, labels are scarce, or existing labels are unreliable. Naturally, these algorithms also have limitations, which are often algorithm-specific (see Section \ref{sec:related_work} for concrete examples). However, many widely used clustering algorithms treat all features (the attributes describing each data point), as equally important, regardless of their actual relevance to the clustering objective. This is particularly problematic in today's heterogeneous data domains, where the relevance of features varies substantially. In such settings, only a subset of features is truly informative. Hence, treating every feature equally can obscure the underlying cluster structure and lead to a failed clustering.

We argue that effective clustering should reflect differences in relevance rather than treating all features uniformly. This aligns with the intuitive notion that even among relevant features, there are likely to be different degrees of relevance. While there has been research on feature weighting techniques to address this issue (\cite{deng2016survey,chakraborty2022}), most existing methods introduce at least one user-defined hyperparameter that can be difficult to tune effectively, or for which there is no objective or universal approach to estimate, limiting their practical applicability. For example, $k$-means-type clustering based on the $t$-distribution (\cite{xiao2025robust}), Entropy-Weighted \texorpdfstring{$k$}{k}-means (\cite{huang2005ewkm}), and LW-\texorpdfstring{$k$}{k}-means (\cite{chakraborty2022}), all require users to specify parameters. These tuning steps are computationally expensive and problematic in unsupervised settings where no labels are available for validation.
A straightforward way to address this issue is to develop a feature weighting clustering algorithm that does not require additional user-defined parameters. This is precisely what we propose in this paper by introducing the Shapley value (\cite{shapley1953value}), in a novel context for feature weighting during clustering. 

Originating in cooperative game theory, the Shapley value provides a principled way to distribute a total contribution among a set of participants based on their marginal contributions across all possible subsets. In our setting, features act as “participants,” and their importance is quantified with respect to the clustering algorithm’s objective function (see Section~\ref{sec:shapley}). This offers a mathematically grounded and axiomatically justified measure of feature relevance. Prior work has applied Shapley values to feature selection for post hoc explainability---where features are either selected or discarded (\cite{li2024unsupervised,xue2024remote})---but, to our knowledge, not to feature weighting during the clustering process itself.

Our contribution is threefold. First, we establish a link between a feature's Shapley value and its dispersion, framing $k$-means clustering as a cooperative game in which features collectively ``explain'' the total clustering cost, and showing that Shapley computation reduces from exponential to polynomial time in this setting. Second, we show that this decomposition leads naturally to an inverse-dispersion weighting scheme, equivalent to replacing the arithmetic mean of feature dispersions optimised by $k$-means with their harmonic mean, providing a guaranteed reduction in clustering cost whenever features differ in relevance. Third, we introduce Shapley Reweighted $k$-means (SHARK), a feature-weighted clustering algorithm free of additional user-defined parameters.

SHARK’s Shapley-based weighting naturally attenuates noisy features and highlights informative ones without manual intervention, and computes weights directly from the Shapley-value decomposition of the $k$-means objective, requiring no additional parameters beyond those in $k$-Means.

Experiments on synthetic and real-world datasets, with and without added noise features, show that SHARK consistently matches or outperforms existing feature-weighted clustering methods.


\section{Related work}
\label{sec:related_work}

This section is divided into two main parts. In Section \ref{sec:clustering} we review clustering algorithms related to our work, beginning with the classical $k$-means and progressing to more modern variants implementing feature weighting. In Section \ref{sec:shapley} we discuss the Shapley value (\cite{shapley1953value}) and its use in \textit{post hoc} feature importance analysis, which inspired our SHARK algorithm (for details, see Section \ref{sec:new_algo}).


\noindent
\subsection{Clustering algorithms}
\label{sec:clustering}

Let \(X = \{x_1, \ldots, x_n\} \) be a data set, where each \( x_i \in X \) is described over \( m \) features. Clustering algorithms aim at producing a clustering \( \mathcal{C} = \{C_1, \ldots, C_k\} \) of \( X \), with each \( C_\ell \in \mathcal{C} \) containing points such that the overall intra-cluster similarity is greater than inter-cluster similarity, by some measure of similarity. There are many approaches that can be used to form such a clustering. Density-based algorithms identify clusters as dense regions of data points separated by low-density areas, making them effective for arbitrary-shaped clusters (\cite{bhattacharjee2021survey,bushra2021comparative}). Hierarchical algorithms build nested clusterings by iteratively merging or splitting clusters based on some objective function. That is, they produce a clustering \( \mathcal{C} \) as well the relationships between the clusters in \( \mathcal{C} \) (\cite{oyewole2023data,shetty2021hierarchical}). Here, we focus on partitional algorithms. These aim at producing a clustering \( \mathcal{C} \) that is a true partition of \(X \). That is, \( C_\ell \cap C_t = \emptyset \) for all \( \ell, t = 1, \ldots, k \) with \( \ell \neq t \), and \( X = \bigcup_{\ell=1}^k C_\ell \).

The $k$-means algorithm (\cite{macqueen1967}) is arguably the most popular partitional clustering algorithm (\cite{ikotun2023k,sinaga2020unsupervised}). It iteratively minimises 
\begin{equation}
    \label{eq:kmeans}
    W(\mathcal{C}, Z) = \sum_{\ell=1}^{k} \sum_{x_i \in C_{\ell}} \sum_{v=1}^m (x_{iv} - z_{\ell v})^2,
\end{equation}
where \( Z = \{z_1, \ldots, z_k\} \) is the set of centroids, with each \( z_\ell \in Z \) being the centroid of a cluster \( C_\ell \). This objective function corresponds to the total within-cluster sum of squared Euclidean distances. The algorithm proceeds by alternating between assigning each \(x_i \in X \) to the cluster of its nearest centroid, and updating each \( z_\ell \in Z \) to be the component-wise mean of all points in \( C_\ell \). This process continues until convergence, typically when the assignments cease to make changes in \( \mathcal{C} \), or the reduction in \( W(\mathcal{C}, Z) \) falls below a pre-defined threshold. Due to its simplicity, scalability, and computational efficiency, $k$-means has become a cornerstone in exploratory data analysis, pattern recognition, and various application domains such as image segmentation, market segmentation, and bioinformatics. 

The above does not imply that $k$-means is without faults. At each iteration, $k$-means minimises \( W(\mathcal{C}, Z) \) using a greedy approach, meaning there is no guarantee the algorithm will reach a global minimum. In fact, the likelihood of reaching global minimum heavily depends on the initial choice of centroids. With this in mind, a number of initialisation algorithms have been devised for $k$-means (see for instance \cite{zhang2024speeding,de2012empirical,kumar2024high,zhang2025structured,harris2022extensive}), with $k$-means++ (\cite{vassilvitskii2006k}) being the most popular. The latter is the default $k$-means initialisation in software packages such as MATLAB, scikit-learn, and R (\cite{scikit-learn,RLanguage,MATLAB}). Hence, this is one of the algorithms we compare against (for details see Section \ref{sec:results}).

A major issue shared by $k$-means and $k$-means++ is the underlying assumption that all features are equally relevant. That is, an irrelevant or noisy feature can contribute to \( W(\mathcal{C}, Z) \) just as much as a perfectly meaningful feature. As a result, the final clustering \( \mathcal{C} \) may be influenced by such uninformative features --- potentially even disproportionally, depending on their prevalence --- and fail to reflect the true cluster structure present in the meaningful features of a data set \( X \). 


There has been considerable research resulting in the development of numerous feature weighting clustering algorithms. However, the vast majority of these methods require the user to specify at least one additional parameter. Identifying suitable parameter values is a non-trivial task that often requires running many experiments, which can become impractical for large data sets. In a recent survey (\cite{hancer2020survey}), \textit{Feature Weight Self-Adjustment $k$-means} (FWSA) (\cite{tsai2008developing}) was the only feature weighting algorithm mentioned that does not require any additional user-defined parameters.

FWSA automatically calculates feature weights during the clustering process by adding a weight adjustment phase to $k$-means. First, it calculates the within-cluster separation of a feature \( v \) as 
\[
    a_v = \sum_{\ell=1}^k \sum_{x_i \in C_\ell} d(x_{iv}, z_{\ell v}),
\]
where \( d(\cdot, \cdot) \) is a distance function. Second, it calculates the between-cluster separation of \( v \) as 
\[
    b_v = \sum_{\ell=1}^k |C_\ell| \cdot d(z_{\ell v}, \bar{z}_v),
\]
where \( \bar{z}_v \) is the centre of \( v \) over all \( x_i \in X \). Using these two measures FWSA evaluates the contribution of \( v \) to the clustering quality at a given iteration \( \tau \), and then sets \( \omega_v^{(\tau+1)} \) to
\[
    \omega_v^{(\tau+1)} = \frac{1}{2}\left(\omega_v^{(\tau)} + \frac{b_v^{(\tau)}/a_v^{(\tau)}}{\sum_{v'=1}^m b_{v'}^{(\tau)}/a_{v'}^{(\tau)}} \right).
\]
Starting with equal weights (typically, \( \omega_v = 1/m \)), the algorithm iterates this rule as part of the clustering process. As a result, it automatically emphasises important features while reducing the influence of less relevant noisy features --- all without requiring any extra user-defined parameter.


The combination of the two distance terms ensures that relevant features are up-weighted and retained, while irrelevant features may receive a weight of exactly zero due to the sparsity-inducing penalty. 

The \textit{Lasso Weighted $k$-means (LW-$k$-means)} \cite{chakraborty2022} algorithm is a sparse clustering method tailored for high-dimensional data. Unlike FWSA and our method (see Section~\ref{sec:new_algo}), LW-$k$-means introduces the parameters $\lambda$, $\alpha$ and $\beta$ (where $\alpha$ can be computed and $\beta$ may be set to a default value). The method enhances classical $k$-means by introducing feature weighting and promoting sparsity through an $\ell_{1}$ regularisation term. It minimises
\begin{align}
W_{\text{LW-$k$-means}}(Z, \omega) 
  &= \sum_{i=1}^{n} \min_{1 \leq \ell \leq k} 
     \left\{ \|x_i - z_\ell\|_{\omega^\beta}^{2} 
     + \lambda \|x_i - z_\ell\|_{|\omega|}^2 \right\} \nonumber \\
  &\quad - \alpha \,\omega^\top 1_m.
\end{align}

The weighted norms are defined by
\[
\|x_i - z_\ell\|_{\omega^\beta}^{2}
    = \sum_{v=1}^m \omega_v^{\beta} (x_{iv} - z_{\ell v})^2,
\]
\[
\|x_i - z_\ell\|_{|\omega|}^{2}
    = \sum_{v=1}^m |\omega_v| (x_{iv} - z_{\ell v})^2.
\]

The combination of the two distance terms ensures that relevant features are up-weighted and retained, while irrelevant features may receive a weight of exactly zero due to the sparsity-inducing penalty. 

To express the weight update, we introduce the cluster membership indicator
\[
u_{i\ell} =
\begin{cases}
1, & \text{if } x_i \in C_\ell, \\
0, & \text{otherwise}.
\end{cases}
\]

The quantity
\[
D_v = \sum_{i=1}^n \sum_{\ell=1}^k u_{i\ell} (x_{iv}-z_{\ell v})^2
\]
is used in the block-coordinate descent update for the weights, which is given by
\[
\omega_v^* =
\begin{cases}
0, & \text{if } D_v = 0, \\
\left[ \frac{1}{\beta} \,
S\!\left( \frac{\alpha}{D_v}, \lambda \right)
\right]^{1/(\beta - 1)}, & \text{otherwise},
\end{cases}
\]
where
\[
S(x, \lambda) =
\begin{cases}
x - \lambda, & \text{if } x > \lambda, \\
x + \lambda, & \text{if } x < -\lambda, \\
0, & \text{otherwise}.
\end{cases}
\]

This retains the computational efficiency of $k$-means, with a per-iteration complexity of \( \mathcal{O}(nkm) \), and guarantees convergence. The original authors further prove strong consistency of the centroids and feature weights under independent and identically distributed (i.i.d.)\ sampling, arguing for both theoretical rigour and practical utility.

Despite its strengths, LW-$k$-means has limitations, some shared with $k$-means. It assumes a known number of clusters \( k \) and is sensitive to initialization, though $k$-means++ seeding improves stability. While the authors propose stability selection to tune \(\lambda\), this process can be computationally intensive --- a drawback that somewhat undermines the method's suitability for high-dimensional data. In cases where all features are informative, the \( \ell_1 \) penalty may undesirably shrink useful weights, leading to suboptimal clustering.

The \textit{$k$-means-type clustering based on the $t$-distribution} (KMTD) \citep{xiao2025robust} algorithm is a recently proposed partitional clustering method designed to improve robustness to noise and outliers. It is motivated by the classical derivation of $k$-means as a small-variance asymptotic approximation of a Gaussian mixture model (GMM), and analogously derives a clustering algorithm from a simplified multivariate $t$-mixture model (TMM). Since the $t$-distribution has heavier tails than the Gaussian, it naturally downweights points that lie far from their assigned cluster centres, thereby providing intrinsic robustness to noisy observations.

KMTD assumes that the data are generated from a mixture of $k$ spherical $t$-distributions sharing a common degrees-of-freedom parameter $\nu$ and a common scale parameter $\sigma > 0$. Under this model, the conditional probability that observation $x_i$ originates from cluster $\ell$ is
\[
    \gamma_{i\ell}
    =
    \frac{
        w_\ell \left( \nu\sigma^2 + \|x_i - z_\ell\|^2 \right)^{-(\nu + m)/2}
    }{
        \sum_{h=1}^k 
        w_h \left( \nu\sigma^2 + \|x_i - z_h\|^2 \right)^{-(\nu + m)/2}
    },
\]
where $w_\ell$ is the mixture proportion of cluster $\ell$ and $z_\ell \in \mathbb{R}^m$ is its centre. The heavy-tailed form of the $t$-density ensures that the contribution of large distances is suppressed relative to the Gaussian case (\cite{peel2000robust}), thereby reducing the impact of outliers on both cluster assignments and centre updates.

Given the responsibilities $\gamma_{i\ell}$ (how strongly each point belongs to each cluster), KMTD updates the mixture proportions by their maximum-likelihood estimates
\[
    w_\ell = \frac{1}{n} \sum_{i=1}^n \gamma_{i\ell},
\]
and updates the cluster centres via a weighted location estimate

\[
    z_\ell^{(\tau+1)} 
    = 
    \frac{
        \sum_{i=1}^n 
        \displaystyle
        \frac{\gamma_{i\ell}^{(\tau)} x_i}{\nu\sigma^2 + \|x_i - z_\ell^{(\tau)}\|^2}
    }{
        \sum_{i=1}^n 
        \displaystyle
        \frac{\gamma_{i\ell}^{(\tau)}}{\nu\sigma^2 + \|x_i - z_\ell^{(\tau)}\|^2}
    }.
\]

This centre update resembles a reweighted mean, where the denominator term $\nu\sigma^2 + \|x_i - z_\ell\|^2$ ensures that contributions from distant observations are significantly downweighted. As a result, cluster centres are less susceptible to being pulled away by extreme values, in contrast to the standard $k$-means update.

The updates for $\gamma_{i\ell}$, $w_\ell$, and $z_\ell$ are iterated until convergence of the soft assignment matrix $\Gamma = (\gamma_{i\ell})$, or until a maximum number of iterations $N_{\max}$ is reached. After convergence, each observation is assigned to the cluster corresponding to its largest posterior weight, yielding a hard partition

\[
    C_\ell = \left\{ x_i : \ell = \arg\max_{h \in [k]} \gamma_{i h} \right\}.
\]

KMTD requires two parameters, $\nu$ and $\sigma$. However, the authors propose the default values of $\nu = 3$ (reflecting a heavy-tailed distribution), and $\sigma = \sigma^\ast$, a data-driven estimate proportional to the overall dispersion of $X$. This simplification reduces its computational cost to $\mathcal{O}(T m k n)$, where $T$ is the number of iterations \citep{xiao2025robust}. 

Empirically, KMTD has demonstrated strong performance in settings where clusters are well-separated but contaminated with noise or atypical observations. Its robustness derives from the fact that outlying points have small posterior weights $\gamma_{i\ell}$ and receive large denominators in the centre updates, thereby exerting minimal influence on $z_\ell$. As with $k$-means and LW-$k$-means, the method assumes a pre-specified number of clusters $k$ and is sensitive to initialisation; in practice, $k$-means++ is often used to provide stable initial centres. Nonetheless, KMTD offers a compelling compromise between the speed of $k$-means and the robustness of full TMM fitting, making it suitable for moderate- to high-dimensional data sets where noisy features or observations may be present.


\subsection{Shapley Values}
\label{sec:shapley}

Shapley values, originally introduced in cooperative game theory (\cite{shapley1953value}), provide a theoretically grounded method for attributing the output of a function to its input variables. In machine learning, they are often used within the context of feature selection to quantify the contribution of each feature to a model's prediction (\cite{yin2022adaptive,al2024genetic,marcilio2021explaining}). 

Let \(V=\{1, \ldots, m\}\) be the feature set and \(f:2^V \to \mathbb{R}\) be a model evaluated on subsets of features. Then, the Shapley value \( \varphi_v \) for each feature \( v \in V \) is defined as

\begin{equation}
\label{eq:shapley}
\varphi_v = \sum_{S \subseteq V \setminus \{v\}} \frac{|S|!(m - |S| - 1)!}{m!} \left[ f(S \cup \{v\}) - f(S) \right].    
\end{equation} \\
The Shapley value satisfies four key axioms that ensure fairness in attribution. \textit{Efficiency} guarantees that the total value, such as a model's prediction, is fully distributed among all features, meaning the sum of the Shapley values equals the model output relative to a baseline. \textit{Symmetry} ensures that features contributing equally across all coalitions (subsets of \(V\)) receive equal Shapley values. The \textit{Null Player} axiom states that a feature that does not affect the output when added to any subset receives a Shapley value of zero. Lastly, \textit{Additivity} means that for any two models, a feature's Shapley value under their sum is equal to the sum of its Shapley values under each model individually.

In practical applications, Shapley values are widely used for explaining individual predictions by attributing the model output to input features in a manner that is consistent and theoretically justified. Despite these advantages, several challenges can limit the direct applicability of Shapley values in practice. Chief among them is time complexity. In most cases, the exact calculation of~\eqref{eq:shapley} requires evaluating the model across all \(2^m\) subsets of features, which becomes unfeasible for high-dimensional data. Another significant issue arises from feature correlation. The Shapley framework assumes that features are independent, and when this assumption is violated, the attribution may become ambiguous or counterintuitive. Additionally, Shapley values reflect how the model uses the data, rather than uncovering causal relationships, which limits their use in settings where causal interpretability is required. Overall, Shapley values offer a robust and flexible approach to interpreting machine learning models, balancing theoretical rigour with practical relevance --- albeit with some limitations that must be addressed in real-world applications.



\section{A Shapley-Theoretic framework}
\label{sec:new_algo}
This section is divided into three main parts. Section \ref{subsec:k_means_shapley} presents a decomposition of $k$-means using Shapley values, offering a theoretical perspective that motivates our SHARK method, introduced in Section \ref{subsec:SHARK}. Finally, in Section \ref{subsec:SHARK_properties} we discuss some theoretical properties of our method.

\subsection{Decomposing $k$-means using Shapley values}
\label{subsec:k_means_shapley}
In this section, we prove that the $k$-means objective can be exactly decomposed as the sum of per-feature Shapley values, leading to a novel axiomatic foundation for unsupervised feature relevance. First, we make a short observation.
\begin{obs}
    \label{obs:kmeans_additive}
    The $k$-means objective is additive. That is, for any feature \(u\) and feature set \(S \subseteq \{1, \ldots, m\} \setminus \{u\}\),
    \begin{align*}
        \sum_{\ell = 1}^k \sum_{x_i \in C_{\ell}} \sum_{v \in S \cup \{u\}} (x_{iv}-z_{\ell v})^2 &- \sum_{\ell = 1}^k \sum_{x_i \in C_{\ell}} \sum_{v \in S} (x_{iv}-z_{\ell v})^2\\
        &= \sum_{\ell = 1}^k \sum_{x_i \in C_{\ell}} (x_{iu}-z_{\ell u})^2.
    \end{align*}
\end{obs}

We are interested in applying the Shapley value framework to quantify the degree of relevance of each feature in $k$-means. Hence, we define a cooperative game in which each feature is a player, and the Shapley characteristic function is the $k$-means objective restricted to a subset \(S\) of features. That is, the function \(f : 2^m \to \mathbb{R}\)
\[
f(S) = \sum_{\ell=1}^k \sum_{x_i \in C_\ell} \sum_{v \in S} (x_{iv} - z_{\ell v})^2.
\]
This choice reflects the portion of the clustering cost attributable to a given subset of features and provides a suitable domain for applying the Shapley value. Let us analyse the impact of this choice.

\begin{theorem}
\label{thm:shapley_kmeans}
The $k$-means algorithm iteratively minimises the sum of the Shapley values over all features, \(\sum_{u=1}^m \varphi_u\).
\end{theorem}
\begin{proof}
Let \(\mathcal{C}\) be the clustering of \(X\). Then, as per Observation \ref{obs:kmeans_additive} the marginal contribution of a feature \(u\) is given by
\[
f\left(S \cup \{u\}\right) - f\left(S\right) = \sum_{\ell=1}^k \sum_{x_i \in C_\ell} (x_{iu} - z_{\ell u})^2,
\]
for any \(S \subseteq \{1, \dots, m\} \setminus \{u\}\). \\

We can then re-write the Shapley value for \(u\),

\begin{align*}
\varphi_u &= \sum_{S \subseteq V \setminus \{u\}} \frac{|S|!(m - |S| - 1)!}{m!} \cdot \left[ f(S \cup \{u\}) - f(S) \right]\\
&= \sum_{S \subseteq V \setminus \{u\}} \frac{|S|!(m - |S| - 1)!}{m!} \sum_{\ell=1}^k \sum_{x_i \in C_\ell} (x_{iu} - z_{\ell u})^2.
\end{align*}\\

Notice that 

\[
\sum_{S \subseteq V \setminus \{u\}} \frac{|S|!(m - |S| - 1)!}{m!} = 1.
\]\\

Hence,

\begin{equation}
\label{eq:shapley_fast}
\varphi_u = \sum_{\ell=1}^k \sum_{x_i \in C_\ell} (x_{iu} - z_{\ell u})^2.    
\end{equation}\\

Substituting the above into the objective iteratively minimised by $k$-means,

\begin{align}
\label{eq:kmeans_phi}
    W(\mathcal{C}, Z) = \sum_{\ell=1}^{k} \sum_{x_i \in C_\ell} \sum_{v=1}^m (x_{iv} - z_{\ell v})^2=\sum_{u=1}^m \varphi_u.
\end{align}\\

Thus, the $k$-means objective is equal to the sum of the Shapley values of all features.
\end{proof}

Theorem~\ref{thm:shapley_kmeans} shows that the \textit{$k$}-means objective function is exactly the sum of Shapley values over features, where the characteristic function \( f(S) \) is the within-cluster dispersion of the data set restricted to feature subset \(S\). Hence, $k$-means can indeed be seen as a cooperative game in which the features act as players, and their combined goal is to ``explain'' the total clustering cost.

This decomposition elevates the role of individual features to strategic contributors in a cooperative game, where their Shapley values reflect average marginal costs over all coalitions. Crucially, it reframes the $k$-means objective not only as an additive cost over features, but as the unique value consistent with fairness axioms from cooperative game theory. This perspective opens the door to principled interpretations of feature relevance, robust weighting schemes based on inverse contribution, and novel algorithms grounded in game-theoretic equilibrium concepts.

\subsection{Shapley reweighted $k$-means}
\label{subsec:SHARK}

This section introduces our Shapley Reweighted $k$-Means (SHARK) algorithm. Our aim is to extend $k$-means by estimating the degree of importance of each feature, \(\omega_v\), and incorporating this into~\eqref{eq:kmeans}. 

First, recall that Theorem \ref{thm:shapley_kmeans} shows \\

\(\varphi_v = \sum_{\ell=1}^k \sum_{x_i \in C_{\ell}} (x_{iv}-z_{\ell v})^2\). \\

Hence, for the SHARK objective function, we can state:

\begin{align}
\label{eq:initial_SHARK}
W(\mathcal{C}, Z, \omega) 
&= \sum_{\ell=1}^{k} \sum_{x_i \in C_\ell} \sum_{v=1}^m \omega_v (x_{iv} - z_{\ell v})^2 \nonumber \\
&= \sum_{v=1}^m \omega_v \varphi_v
\end{align}\\

In this context, the Shapley value \(\varphi_v\) of a feature \(v\) reflects the increase in cost of~\eqref{eq:kmeans} when $v$ is added to a subset of features. That is, the higher the value of \(\varphi_v\), the more \(v\) worsens~\eqref{eq:kmeans}. Hence, it is natural to penalise such features by setting \(\omega_v \propto \varphi_v^{-1}\). It is worth noting that, at each iteration, the Shapley values $\varphi_v$ are computed with respect to the current cluster assignments $\mathcal{C}$ and centroids $Z$. Consequently, the feature weights $\omega_v$ reflect feature relevance relative to the current partition rather than the globally optimal clustering. This is analogous to the greedy nature of $k$-means itself, where each assignment and centroid update step optimises locally. The iterative reweighting nevertheless converges to a stable solution, as the weights and cluster assignments are updated jointly until $\mathcal{C}$ no longer changes.

We normalise the weight so that \\

\(\sum_{v=1}^m \omega_v=1\), \\

ensuring the total contribution of all features remains constant and avoiding degenerate solutions. 

Thus,

\begin{equation}
    \label{eq:SHARK_weight}
    \omega_v = \frac{\varphi_v^{-1}}{\sum_{u=1}^m \varphi_u^{-1}}.
\end{equation}\\

We can now further develop the SHARK objective by substituting~\eqref{eq:SHARK_weight} into~\eqref{eq:initial_SHARK},

\begin{align}
\label{eq:SHARK}
W(\mathcal{C}, Z, \omega) 
&= \sum_{v=1}^m \omega_v \varphi_v \nonumber \\
&= \sum_{v=1}^m \frac{\varphi_v^{-1}}{\sum_{u=1}^m \varphi_u^{-1}} \cdot \varphi_v \nonumber \\
&= \sum_{v=1}^m \frac{1}{\sum_{u=1}^m \varphi_u^{-1}} \nonumber \\
&= \frac{m}{\sum_{u=1}^m \varphi_u^{-1}}.
\end{align}

That is, SHARK minimises the harmonic mean of the Shapley values. Algorithm~\ref{alg:SHARK} outlines the specific steps SHARK uses to iteratively minimise \eqref{eq:SHARK}. It calculates feature weights at each iteration, effectively adding an extra step in relation to $k$-means. These weights guide the clustering process by emphasising features that contribute the least to the cost. These are the features most consistent with the underlying cluster structure in the data.

\begin{algorithm}[H]
\caption{Shapley Reweighted $k$-means (SHARK)}
\label{alg:SHARK}
\begin{algorithmic}[1]
\REQUIRE Data set \(X =\{x_1, \ldots, x_n\}\) with \(x_i \in \mathbb{R}^m\), number of clusters $k$.
\OUTPUT Clustering \(\mathcal{C}=\{C_1, \ldots, C_k\}\), feature weights $\omega \in \mathbb{R}^m$.
\STATE Set $\omega_v = \frac{1}{m}$ for all $v = 1, \ldots, m$.
\STATE Select \(k\) distinct data points from \(X\) uniformly at random, and copy their values to \(z_1, \ldots, z_k\).
\REPEAT
    \STATE Assign each \(x_i \in X\) to the cluster \(C_\ell\), where
    \[
    \ell = \arg\min_{t=1}^k \sum_{v=1}^m \omega_v (x_{iv} - z_{tv})^2,
    \]
    breaking ties by choosing the smallest such index \(t\).
    \STATE Update each centroid \(z_\ell \in Z\) to the component-wise mean of the points in \(C_\ell\).
    \STATE Update \(\omega_v\) for all \(v = 1, \ldots, m\) using Equations~\eqref{eq:SHARK_weight} and~\eqref{eq:shapley_fast}.
\UNTIL{the clustering \(\mathcal{C}=\{C_1, \ldots, C_k\}\) converges}
\RETURN $\mathcal{C}$, $\omega$
\end{algorithmic}
\end{algorithm}

\subsection{SHARK Properties}
\label{subsec:SHARK_properties}

We begin by proving that SHARK converges in a finite number of iterations.
\begin{theorem}
SHARK converges in a finite number of iterations.
\end{theorem}
\begin{proof}
At each iteration, the three update steps do not increase the SHARK objective~\eqref{eq:SHARK}. The assignment step does not increase it by 
the same argument as $k$-means: each point is assigned to its nearest centroid under the current weights. The centroid update does not increase it as each centroid is updated to the component-wise mean of its cluster, which minimises the within-cluster sum of squared distances. Finally, the weight update does not increase it as~\eqref{eq:SHARK_weight} yields the weights that minimise~\eqref{eq:SHARK} for the current partition. Since the objective is bounded below by zero and is non-increasing, and the number of distinct partitions is finite, SHARK must terminate in a finite number of iterations.
\end{proof}

Note that, given \\

\(\sum_{v=1}^m \omega_v= 1\), \\

we have that \\

\(\sum_{v=1}^m \varphi_v > \sum_{v_1}^m \omega_v \varphi_v\), for \(m>1\). \\

That is, the $k$-means objective~\eqref{eq:kmeans_phi} is always higher than that of SHARK~\eqref{eq:initial_SHARK} if the data set has more than one feature. Hence, these are not comparable leading us to the following observation.

\begin{obs}
    \label{obs:comparable_kmeans}
    The $k$-means objective is comparable to that of SHARK if we divide the former by $m$. This is the case as,
    \[
     \frac{\sum_{v=1}^m \varphi_v}{m}= \frac{1}{m} \sum_{v=1}^m \varphi_v = \sum_{v=1}^m \frac{1}{m}\varphi_v.
    \]
\end{obs}

\begin{figure}[hpt!]
    \centering
    \includegraphics[width=0.85\linewidth]{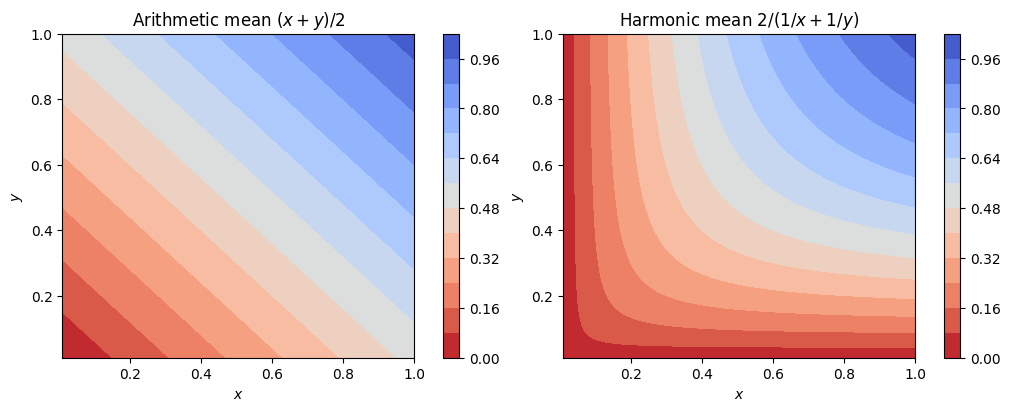}
    \caption{
        \textbf{Comparison between the arithmetic and harmonic mean of two variables \(x\) and \(y\).} }
    \label{fig:arith_harmonic_comparison}
\end{figure}

We now have that the SHARK objective is the harmonic mean over the set \(\{\varphi_1, \ldots, \varphi_m\}\) (see Section \ref{subsec:SHARK}), while the comparable $k$-means is the arithmetic mean over the same set (see Observation \ref{obs:comparable_kmeans}). 
Recall that for any set with cardinality bigger than one (i.e. a set containing different values) its harmonic mean is always less than its arithmetic mean (see Figure \ref{fig:arith_harmonic_comparison} for a visual representation). Hence, if not all values of \(\phi\) are the same, then the SHARK objective is always less than that of $k$-means for the same clustering.

\begin{theorem}
    Let \(m>1\), and suppose that not all values of \(\varphi_1, \ldots, \varphi_m\) are equal. Then, given the same set of centroids \(Z\), SHARK will produce a clustering with a lower cost than that of the comparable $k$-means.
\end{theorem}
\begin{proof}
We have that $k$-means and SHARK minimise the arithmetic and harmonic mean of \(\varphi_1, \ldots, \varphi_m\), respectively. Since \(m > 1\) and not all \(\varphi_v\) are equal, the arithmetic mean is strictly greater than the harmonic mean. Hence, when given the same centroids \(Z\), the clustering obtained by SHARK leads to a strictly lower objective value than the clustering obtained by $k$-means.
\end{proof}

The above implies that, for identical initial centroids, the SHARK's objective is already lower than that of the comparable $k$-means after just one iteration. This potentially leads to a faster descent in the objective. 

We now can express the difference between the objectives as
\[
\varepsilon = \frac{1}{m} \sum_{v=1}^m \varphi_v -  \sum_{v=1}^m \omega_v \varphi_v,
\]
and proceed to identify its lower bound.

\begin{theorem}
\label{thm:SHARK_lowerbound}
Let $\varphi_1, \ldots, \varphi_m > 0$ with $m>1$. Then,
\[
\varepsilon \ \ge \ \frac{(\max_v \varphi_v - \min_v \varphi_v)^2}{2(\max_v \varphi_v + \min_v \varphi_v)}.
\]
In particular, if either \(\max_v \varphi_v\) or \(\min_v \varphi_v\) stay fixed, then this lower bound increases with \(\max_v \varphi_v - \min_v \varphi_v\).
\end{theorem}

\begin{proof}
Let $S$ be a set containing positive values. Recall the inequality,
\[
A - H \ \ge \ \frac{(\max S - \min S)^2}{2(\max S + \min S)},
\]
where $A$ and $H$ are the arithmetic and harmonic means of the elements in $S$ (\cite{sharma2008some}). Since $\varepsilon = A - H$, the inequality follows by setting $S=\{\varphi_1, \ldots, \varphi_m\}$.

Let us now address the second part of the theorem. For fixed $\min S>0$ and $\max S$ replaced by a larger value, the numerator $(\max S - \min S)^2$ increases while the denominator $\max S + \min S$ also increases but less so proportionally. Hence, the bound, increases. For fixed $\max S>0$ and $\min S$ replaced by a smaller value, the numerator $(\max S - \min S)^2$ increases while the denominator $\max S + \min S$ decreases. Hence, the bound increases.
\end{proof}

\begin{cor}
    Let $\varphi_{m+1}>0$ be a new feature. Then, the lower bound for \(\varepsilon\) does not decrease, and it is strictly larger if \(\varphi_{m+1}\notin[\min_{v=1}^m \varphi_v,\max_{v=1}^m \varphi_v]\).
\end{cor}
\begin{proof}
    Let \(\varphi_{m+1} \in [\min_{v=1}^m \varphi_v,\max_{v=1}^m \varphi_v]\). Then, there is no change in the lower bound in Theorem \ref{thm:SHARK_lowerbound}. If \(\varphi_{m+1}\notin[\min_{v=1}^m \varphi_v,\max_{v=1}^m \varphi_v]\), then either \(\varphi_{m+1} < \min_{v=1}^m \varphi_v\) or \(\varphi_{m+1} > \max_{v=1}^m \varphi_v\). In either case, the difference \(\min_{v=1}^{m+1} \varphi_v - \max_{v=1}^{m+1} \varphi_v\) increases with one extreme fixed. Hence, the bound increases by Theorem \ref{thm:SHARK_lowerbound}.
\end{proof}

SHARK replaces the arithmetic mean (used implicitly by k-means) with the harmonic mean, which naturally suppresses noisy or weakly relevant dimensions. As a result, SHARK always performs at least as well as $k$-means, and often much better when some features are uninformative.

\subsection{Time complexity of SHARK}
\label{subsec:SHARK_complexity}
The per-iteration complexity of SHARK follows closely that of $k$-means. 
Recall from Observation~\ref{obs:kmeans_additive} and Eq.~\eqref{eq:shapley_fast} that the Shapley value of each feature is obtained in closed form as
\[
\varphi_v = \sum_{\ell=1}^k \sum_{x_i \in C_\ell} (x_{iv} - z_{\ell v})^2,
\]
that is, as the feature-wise contribution to the $k$-means objective. 
Computing all $\varphi_v$ therefore requires no combinatorial enumeration and adds no extra asymptotic cost beyond that of evaluating the weighted distances already used in the assignment step. 
Thus, each iteration of SHARK incurs the same $\mathcal{O}(nkm)$ cost as $k$-means, with only an additional $\mathcal{O}(m)$ normalisation step for the weights. Consequently, SHARK preserves the scalability of $k$-means while introducing a negligible overhead for feature reweighting.

\section{Experimental setting}

We validate our clustering method, SHARK, by comparing it against existing approaches on a combination of real-world and synthetic data sets, with and without adding extra noise features to these.

\subsection{Data sets and measuring cluster recovery}
\label{sec:datasets_and_measures}

We initially generated 600 synthetic data sets under 12 configurations, with each data set containing spherical Gaussian clusters. Each cluster had a diagonal covariance matrix with variance \(\sigma^2\) drawn from the uniform distribution \(\mathcal{U}(0.5, 1.5)\). Hence, our data sets contain clusters that differ in spread, leading to non-uniform overlap and varying separation between clusters. Cluster centroids were sampled independently from the multivariate normal distribution \(\mathcal{N}(0,I)\). The size of each cluster was sampled uniformly at random, with a minimum of 20 elements. We generated 50 data sets for each of the configurations below.

\begin{enumerate}[label=(\roman*)]
    \item 1000x10-3, 1000x10-5, and 1000x10-10: each data set contains 1,000 data points described over 10 features and partitioned into 3, 5, and 10 clusters, respectively.
    \item 2000x20-5, 2000x20-10, and 2000x20-20: each data set contains 2,000 data points described over 20 features and partitioned into 5, 10, and 20 clusters, respectively.
    \item 2000x30-5, 2000x30-10, and 2000x30-20: each data set contains 2,000 data points described over 30 features and partitioned into 5, 10, and 20 clusters.
    \item 5000x50-10, 5000x50-20, and 5000x50-50: each data set contains 5,000 data points described over 50 features and partitioned into 10, 20, and 50 clusters.
\end{enumerate}

To simulate the presence of irrelevant features, we appended 50\% additional noise features to each data set. These noise features were sampled independently from a uniform distribution. As a result, each configuration was evaluated in both clean and noisy settings, effectively doubling the number of synthetic data sets used in the experiments to a total of 1,200. In terms of notation, we indicate data sets with added noise features with the acronym NF. For instance, 2000x30-5 +15NF is the configuration with data sets containing 2,000 data points initially described over 30 features and partitioned into 5 clusters, to which we added 15 extra noise features, resulting in a total of 45 features.

We applied the range normalisation to all data sets above. That is,
\[
x_{iv} = \frac{x_{iv} - \bar{x}_v}{\max (x_v) - \min (x_v)},
\]
where \(\bar{x}_v\), max(\(x_v\)), min(\(x_v\)) are the average, maximum, and minimum of feature \(v\) over all data points \(x_i \in X\), respectively. We opted for range normalisation rather than the more popular $z$-score, because the latter tends to favour unimodal distributions. For example, consider a data set with two features of equal range, a unimodal $v_1$ and a bimodal $v_2$. The standard deviation of $v_2$ is typically higher than that of $v_1$, which leads to lower normalised values under the $z$-score. This effectively penalises multimodal features. However, in the specific case of FWSA, range normalisation led to substantially lower performance. This is likely due to the fact FWSA adjusts feature weights based on the ratio of between- to within-cluster dispersions. Hence, for this particular algorithm we used the $z$-score.

In addition, a small number of real-world data sets were selected from the UCI Machine Learning Repository \citep{uci2019}, the limit being constrained by the need for the LW$k$-means algorithm, a key benchmark, to identify its optimal \(\lambda\) hyperparameter (see section~\ref{subsec:alg_config}). 



Given that ground-truth labels are available for each data set, we measured cluster recovery using the Adjusted Rand Index (ARI) (\cite{hubert1985comparing}). ARI quantifies the agreement between the clustering produced by each algorithm and the true labels, while correcting for chance agreement. This makes it particularly suitable for evaluating clustering performance. 

Section~\ref{sec:results} also shows time comparisons for all algorithms we experiment with. These experiments were executed on a Core i7-9700K CPU (8 Cores, each running at 3.6 GHz)  with 80GB of RAM.

\subsection{Algorithm configuration and execution}
\label{subsec:alg_config}

This paper introduces the feature weighting clustering algorithm SHARK (see Section \ref{sec:new_algo}), which requires no additional user-defined parameter beyond those already present in $k$-means. That is, the data set and the desired number of clusters. This is a significant advantage, as the estimation of optimum parameters is a non-trivial task, often requiring extensive tuning procedures that lack universal or objective guidelines. Despite the practical importance of parameter-free methods, our review of the literature found that comparable algorithms are rare. To our knowledge, FWSA is the only existing feature weighting clustering algorithm that also avoids the introduction of extra parameters. In contrast, other popular methods such as LW-$k$-means and KMTD do require one or more tunable parameters.

In the case of LW-$k$-means, the parameter \(\lambda\) plays an important role. If set too high, it can result in failure to recover clusters. However, for the other parameter, \(\beta\), its authors provide a default value. To tune \(\lambda\) the original authors propose a stability-based selection method. In this, they suggest evaluating 1,000 evenly spaced values in the interval [0,1]. For each test \(\lambda\), the algorithm should be run 50 times on two random samples of \(X\), and clusterings compared. The most stable \(\lambda\) is then selected. However, this process requires up to \(1,000 \times 50 \times 2=100,000\) runs of LW-$K$-means, which is computationally impractical in many settings. Hence, we only apply this parameter tuning method to small real-world data sets. In large synthetic data sets we ran LW-$k$-means with \( \lambda= 0.005\), the value used by the original authors for their experiments on synthetic data sets. In cases this value failed to produce a valid clustering, we progressively decreased \(\lambda\) by an order of magnitude, until the algorithm returned a successful clustering. We ran all algorithms 25 times on each data set, and report the average ARI and respective standard deviation. We supplied each algorithm with the correct number of clusters, $k$. Any clustering containing less than $k$ clusters was considered a failure and discarded. 


KMTD introduces two parameters. These are the degrees of freedom, $\nu$, and the scale parameter, $\sigma$. In our experiments we have used the default values recommended by the original authors ($\nu = 3$ and $\sigma = \sigma^\ast$, a data-driven dispersion estimate). KMTD is sensitive to initialisation, and its authors explicitly advocate the use of multiple random initialisations to improve stability. Accordingly, for each data set we ran KMTD 25 times, each with independent random seeds and a default replication budget of up to 25 internal restarts. Each run produces a hard clustering obtained by assigning each point to the component with the highest posterior probability. As with the other algorithms, any clustering containing fewer than $k$ non-empty clusters was treated as a failure and discarded. Since KMTD does not optimise a direct objective function comparable across runs (such as the LW-$k$-means criterion), we followed the practice used in the original paper and selected among the valid runs the clustering with the highest Average Silhouette Width \citep{Rousseeuw1987}. The final ARI statistics reported for KMTD are therefore computed over the valid runs across all data sets, in direct alignment with the evaluation procedure used for all algorithms we compare.

\section{Results and discussion}
\label{sec:results}

This section presents a detailed analysis of our experimental findings, which empirically validate our SHARK algorithm. Table~\ref{tab:Synthetic_No_Noise} shows results for our experiments on synthetic data sets without injected noise features. Given that in these data sets all features have approximately the same degree of relevance, this (rare) scenario does not favour any algorithm with a feature weighting scheme. Hence, $k$-means++ does well but SHARK is still competitive or better than all algorithms we compare against. This is clear when one notices that SHARK attained the best or second-best average Adjusted Rand Index (ARI) across almost all configurations.
We also note that SHARK attained these results while requiring no parameter tuning and being much faster to converge than its competitors (see Tables~\ref{tab:running_time} and~\ref{tab:running_time_normalised}).

While SHARK generally achieves high clustering performance, a few test configurations exhibit a relative reduction in its effectiveness in terms of cluster recovery (measured with ARI). These are particularly visible in settings without injected noise features and high feature dimensionality, such as 5000x50-50k. In such cases the number of features makes the clusters rather well described, and feature weighting provides no advantage (as all features have approximately the same degree of relevance). Despite this, SHARK still maintains competitive or superior performance overall, and the observed reductions are modest rather than indicative of failure. The findings suggest that while SHARK is broadly robust, its relative benefit may diminish in extremely fine-grained clustering tasks with high dimensionality and no distinction in the relevance of features.

Our experiments reveal that SHARK provides its strongest improvements in settings where the feature relevance structure is heterogeneous. In particular, SHARK consistently yields substantial gains when the data contain redundant or noisy features, when only a subset of dimensions determines the cluster structure, or when high dimensionality amplifies irrelevant variation. In such cases, the Shapley-based reweighting mechanism effectively suppresses uninformative dimensions, improving the separability of the underlying clusters. In contrast, SHARK offers only modest improvements (and may perform comparably to standard $k$-means), in extremely high-dimensional settings without injected noise or redundancy. When all features contribute uniformly to cluster formation, the feature-dispersion scores $\phi$ become nearly homogeneous, and the harmonic-mean objective reduces to behaviour close to that of $k$-means. Under these conditions, no feature-weighting method can exploit differences between dimensions, since the feature space provides no structural signal to reweight.

\begin{table*}[htb!]\small

\begin{center}
\scalebox{0.9}{
\makebox[\textwidth][c]{%
\begin{tabular}{lcccccccc}

\toprule
Experiment & $k$-means++ & FWSA & LW$k$-means & KMTD & SHARK\\
\midrule
1000x10-3k & $0.867 \pm 0.090$ & $0.772 \pm 0.248$ & $\mathbf{0.880} \pm 0.083$ & $0.843 \pm 0.133$ & $\mathbf{0.880} \pm 0.082$\\
1000x10-5k & $0.783 \pm 0.095$ & $0.483 \pm 0.351$ & $\mathbf{0.802} \pm 0.091$ & $0.741 \pm 0.122$ & $\mathbf{0.802} \pm 0.089$\\
1000x10-10k & $0.674 \pm 0.078$& $0.297 \pm 0.302$ & $\mathbf{0.687} \pm 0.075$ & $0.589 \pm 0.075$ & $\mathbf{0.687} \pm 0.077$\\
2000x20-5k & $0.973 \pm 0.021$ & $0.969 \pm 0.025$ & $\mathbf{0.976} \pm 0.019$ & $0.973 \pm 0.021$ & $\mathbf{0.976} \pm 0.019$\\
2000x20-10k & $0.953 \pm 0.018$ & $0.952 \pm 0.020$ & $0.940 \pm 0.129$ & $0.871 \pm 0.058$ & $\mathbf{0.958} \pm 0.016$\\
2000x20-20k & $\textbf{0.914} \pm 0.022$ & $0.909 \pm 0.023$ & $0.905 \pm 0.030$ & $0.284 \pm 0.063$ & $0.911 \pm 0.026$ \\
2000x30-5k & $0.997 \pm 0.003$ & $0.997 \pm 0.004$ & $0.647 \pm 0.115$ & $0.997 \pm 0.003$ & $\mathbf{0.998} \pm 0.003$\\
2000x30-10k & $0.994 \pm 0.004$ & $0.994 \pm 0.004$ & $0.970 \pm 0.088$ & $0.290 \pm 0.075$ & $\mathbf{0.995} \pm 0.004$\\
2000x30-20k & $\textbf{0.983} \pm 0.019$ & $0.958 \pm 0.032$ & $0.794 \pm 0.061$ & $0.263 \pm 0.093$ & $0.977 \pm 0.026$ \\
5000x50-10k & $\mathbf{1.000} \pm 0.000$ & $\mathbf{1.000} \pm 0.000$ & $0.995 \pm 0.025$ & $0.911 \pm 0.040$ & $0.997 \pm 0.019$ \\
5000x50-20k & $\mathbf{1.000} \pm 0.000$ & $0.949 \pm 0.027$ & $0.901 \pm 0.051$ & $0.705 \pm 0.096$ & $0.945 \pm 0.021$ \\
5000x50-50k & $\textbf{0.969} \pm 0.013$ & $0.919 \pm 0.017$ & $0.887 \pm 0.029$ & $0.474 \pm 0.074$ & $0.922 \pm 0.020$ \\
\midrule
Mean Relative Rank & $1.9$ & $3.3$ & $3.1$ & $4.3$ & $\mathbf{1.6}$\\
\bottomrule
\end{tabular}}
} 
\end{center}
\caption{\textbf{Comparison between algorithms using synthetic data, with no injected noise features.} In each case, the best ARI was selected from the best criterion output for the algorithm of 25 experiments for each of 50 datasets per configuration. Results are presented as ARI $\pm$ standard deviation, with the best performer highlighted in bold.}
\label{tab:Synthetic_No_Noise}
\end{table*}

Table~\ref{tab:Synthetic_With_Noise} shows the results for our experiments on synthetic data sets to which we injected noise features. In these experiments, SHARK achieved the highest mean relative rank (1.0), indicating its effectiveness in identifying and down-weighting irrelevant features. In contrast, $k$-means++ exhibited severe degradation in ARI, and LW-$k$-means showed higher variance and sensitivity to the regularization parameter. Although FWSA is also parameter-free, SHARK outperformed it in both average accuracy and consistency, particularly in high-dimensional and noisy settings. These results empirically validate the theoretical formulation of SHARK, specifically its minimization of the harmonic mean of feature-wise Shapley values. This guarantees a clustering cost no worse than that of $k$-means under uniform weighting, and typically much better when irrelevant dimensions are present.

\begin{table*}[htb!]\small
\begin{center}
\scalebox{0.9}{
\makebox[\textwidth][c]{%
\begin{tabular}{lcccccccc}

\toprule
Experiment & $k$-means++ & FWSA & LW$k$-means & KMTD & SHARK\\
\midrule
1000x10-3k+5NF &$0.308 \pm 0.346$ & $0.774 \pm 0.249$ & $0.841 \pm 0.144$ & $0.314 \pm 0.334$ & $\mathbf{0.880} \pm 0.082$ \\
1000x10-5k+5NF &$0.061 \pm 0.116$ & $0.496 \pm 0.345$ & $\mathbf{0.800} \pm 0.093$ & $0.072 \pm 0.115$ & $\mathbf{0.800} \pm 0.091$ \\
1000x10-10k+5NF &$0.017 \pm 0.020$ & $0.216 \pm 0.274$ & $0.678 \pm 0.082$ & $0.008 \pm 0.007$ & $\mathbf{0.680} \pm 0.082$ \\
2000x20-5k+10NF & $0.729 \pm 0.279$ & $0.969 \pm 0.025$ & $\mathbf{0.976} \pm 0.019$ & $0.750 \pm 0.259$ & $\mathbf{0.976} \pm 0.019$ \\
2000x20-10k+10NF &$0.074 \pm 0.056$ & $0.951 \pm 0.021$ & $0.921 \pm 0.060$ & $0.045 \pm 0.051$ & $\mathbf{0.958} \pm 0.017$ \\
2000x20-20k+10NF &$0.017 \pm 0.015$ & $0.912 \pm 0.023$ & $0.891 \pm 0.037$ & $0.002 \pm 0.001$ & $\mathbf{0.917} \pm 0.021$  \\
2000x30-5k+15NF & $0.986 \pm 0.073$ & $0.997 \pm 0.004$ & $0.378 \pm 0.332$ & $0.996 \pm 0.004$ & $\mathbf{0.998} \pm 0.003$\\
2000x30-10k+15NF & $0.655 \pm 0.265$ & $0.994 \pm 0.004$ & $0.936 \pm 0.143$ & $0.376 \pm 0.200$ & $\mathbf{0.995} \pm 0.004$\\
2000x30-20k+15NF &$0.083 \pm 0.045$ & $0.966 \pm 0.029$ & $0.870 \pm 0.042$ & $0.002 \pm 0.003$ & $\mathbf{0.986} \pm 0.012$\\
5000x50-10k+25NF & $\mathbf{1.000} \pm 0.000$ & $\mathbf{1.000} \pm 0.000$ & $0.790 \pm 0.066$ & $0.702 \pm 0.095$ & $\mathbf{1.000} \pm 0.000$ \\
5000x50-20k+25NF & $0.552 \pm 0.280$ & $0.962 \pm 0.030$ & $0.907 \pm 0.048$ & $0.019 \pm 0.018$ & $\mathbf{0.989} \pm 0.023$ \\
5000x50-50k+25NF & $0.034 \pm 0.017$ & $0.928 \pm 0.018$ & $0.922 \pm 0.031$ & $0.021 \pm 0.004$ & $\mathbf{0.969} \pm 0.015$ \\
\midrule
Mean Relative Rank & $4.0$ & $2.3$ & $2.8$ & 4.6 & $\mathbf{1.0}$\\
\bottomrule
\end{tabular}}
} 
\end{center}
\caption{\textbf{Experiments comparing algorithms using synthetic data, with injected noise features.} In each case, the best ARI was selected from the best criterion output of 25 experiments for each of 50 datasets per configuration. Results are presented as ARI $\pm$ standard deviation, with the best performer highlighted in bold.}
\label{tab:Synthetic_With_Noise}
\end{table*}

Table~\ref{tab:Real_World_No_Noise} shows the results of our experiments on real-world data sets from the UCI Machine Learning repository \citep{uci2019}. We were forced to experiment with a low number of small data sets because otherwise LW-$k$-means would simply take too long to identify the correct parameter values (for details, see Section~\ref{subsec:alg_config}). In these experiments, SHARK achieves the best mean relative rank of $1.9$, ahead of FWSA ($2.4$), $k$-means++ and KMTD (both $3.0$), and LW-$k$-means ($4.4$). SHARK attains the best ARI on three of the seven data sets (Blood Transfusion, Iris, and New Thyroid), and ranks among the top two performers on five. On the remaining data sets, FWSA performs strongly on Glass and Iris, while KMTD achieves strong results on Telugu Vowels and Wine. These results are consistent with the behaviour observed in the synthetic experiments without injected noise features: when all features contribute approximately equally to the cluster structure, the advantage conferred by Shapley-based reweighting diminishes. Nonetheless, SHARK's consistently strong mean relative rank across these heterogeneous real-world data sets confirms that its theoretical properties translate reliably into practical performance.

\begin{table*}[htb!]\small
\begin{center}
\scalebox{0.9}{
\makebox[\textwidth][c]{
\begin{tabular}{lcccccc}
\toprule
& $k$-means++ & FWSA & LW$k$-means & KMTD & SHARK \\
\midrule
Blood Transfusion&-0.006&0.041&0.034 & 0.030 &\textbf{0.065}\\
Glass&0.170&\textbf{0.240}&0.141 & 0.123 & $0.218$ \\
Iris&0.716&\textbf{0.886}&0.470 & 0.654 & $\textbf{0.886}$ \\
New Thyroid & 0.583 & 0.205 & 0.194 & 0.117 & \textbf{0.641}\\
Tulugu Vowels&0.380&0.357&0.280 & \textbf{ 0.897} &0.411\\
Wine & \textbf{0.897} & 0.633 & 0.474 & \textbf{0.897} & 0.822\\
Yeast&0.157&0.171&0.139 & \textbf{0.173} &0.160\\
\midrule
Mean Relative Rank & 3.0 & 2.4 & 4.4 & 3.0 & \textbf{1.9} \\
\bottomrule
\end{tabular}}
} 
\end{center}
\caption{\textbf{Experiments comparing algorithms using real world data sets.} In each case, the best ARI was selected from the best criterion output for the algorithm of 25 experiments per dataset. Results are presented as ARI, with the best performer highlighted in bold.}
\label{tab:Real_World_No_Noise}
\end{table*}

\begin{figure}[p!]
    \centering
    \includegraphics[width=1\linewidth]{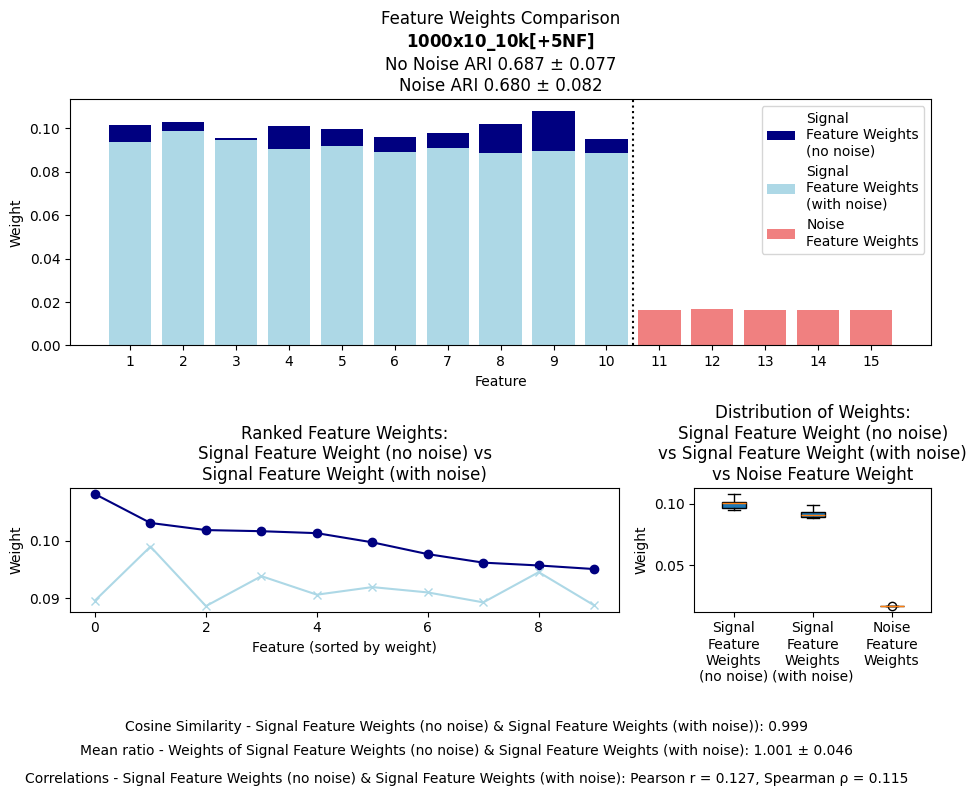}
    \includegraphics[width=1\linewidth]{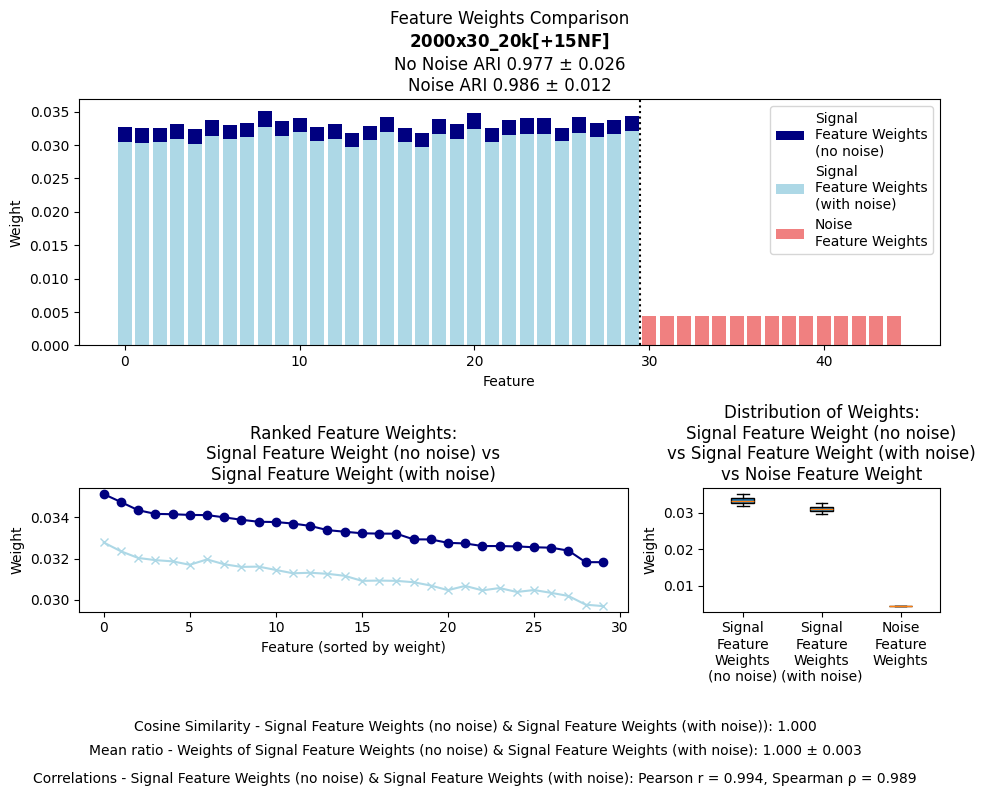}
    \includegraphics[width=1\linewidth]{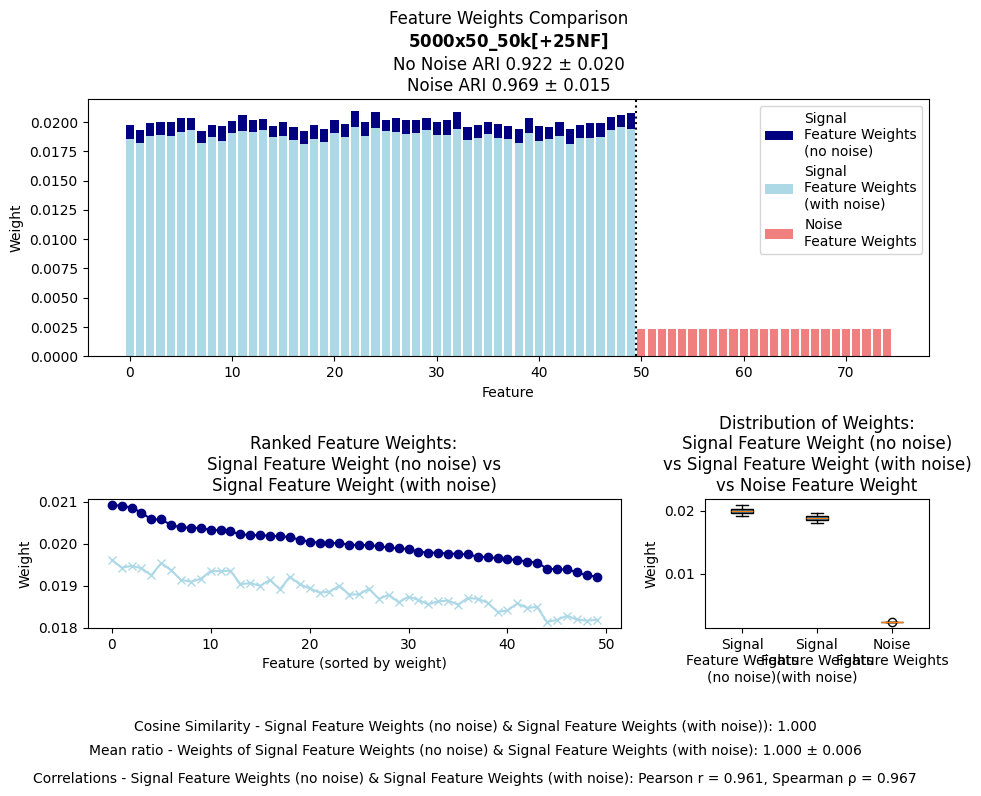}
    \caption{
        Feature weight analysis for different data set configurations, with and without injected noisy features.
    }
    \label{fig:weight_analysis}
\end{figure}

Figure~\ref{fig:weight_analysis} provides a detailed comparison of the feature weights produced by SHARK in both noise-free and noisy settings. SHARK exhibits exceptionally high stability in its estimated weights. In the larger-scale experiments (e.g.\ 5000$\times$50--50$k$ and 2000$\times$30--20$k$), the weights assigned to the informative features remain nearly identical when noise features are injected, with cosine similarities of 1.0 and Pearson correlations exceeding 0.96. Noise features are consistently assigned weights close to zero, producing a clear separation between relevant and irrelevant dimensions. These results show that SHARK’s Shapley-based weighting effectively suppresses uninformative features while preserving the relative importance of the true signal dimensions, even in high-dimensional settings or when a substantial number of noise features are added. Moreover, the inclusion of noise often leads to improved clustering accuracy (as measured by ARI), indicating that SHARK is capable of eliminating irrelevant variation without distorting the underlying cluster structure.

The only configuration in which weight stability deteriorates is the most challenging scenario (1000$\times$10--10$k$), where the small number of informative features and large number of clusters already limit cluster recoverability. In this case, ARI values are low for both noise-free and noisy data, and SHARK’s ranking of signal features becomes less stable. Nonetheless, the algorithm continues to assign markedly lower weights to the injected noise dimensions, demonstrating that the reduction in weight stability reflects the difficulty of the clustering task rather than a failure of the weighting mechanism. Taken together, these findings confirm that SHARK provides robust and interpretable feature weights in settings where meaningful structure is recoverable, while remaining resilient to the presence of noise and irrelevant features.

Finally, Table~\ref{tab:running_time} reports the running times for all algorithms under comparison (hardware details are provided in Section~\ref{sec:datasets_and_measures}). These results show that SHARK introduces only a modest overhead relative to $k$-means++, with running times remaining within the same order of magnitude across all configurations. In contrast, the remaining competitors incur substantially higher computational costs, which is particularly when we normalise these results (see Table~\ref{tab:running_time}). FWSA and L\textnormal{-}W$k$-means scale poorly with both dimensionality and the number of clusters, with L\textnormal{-}W$k$-means in particular exhibiting multi-hour runtimes in several of the larger experiments. KMTD incurs significant computational overhead, with runtimes increasing sharply as either $k$ or the number of noise features grows.

Across all experimental settings, SHARK remains within the same order of magnitude as $k$-means++ in runtime, and is substantially faster than all other competitors. We note that on the largest configurations SHARK can be up to an order of magnitude slower than $k$-means++ itself, which is unsurprising given the additional per-iteration weight update step. Nevertheless, SHARK remains far more efficient than FWSA, LW-$k$-means, and KMTD, while outperforming or matching them in clustering accuracy. This confirms that SHARK achieves its improvements without compromising computational efficiency, preserving the scalability that makes $k$-means attractive while avoiding the substantial overheads associated with more complex alternatives.

\subsection{Case study}

In this section we demonstrate how SHARK can be applied in a realistic, applied setting and to provide qualitative insight into its behaviour beyond aggregate benchmark results. Rather than focusing on large-scale performance comparisons (we have shown these in our previous results), this case study examines a single real-world data set in detail, highlighting how the Shapley-based feature weights evolve, and how they affect the resulting clustering.

This case study uses a malware data set, which consists of 2,000 drive-by-download malware samples collected via the Cuckoo Sandbox 2.0.6 \citep{de2021identifying}. Each sample is represented by 67 features, combining binary indicators and numerical counts extracted from both behavioural and static analyses (e.g. API calls, registry access, network activity, and PE characteristics). All samples were confirmed as malicious by VirusSign using multiple anti-virus engines.

Here we follow the same experimental methodology used in the main evaluation. All algorithms were run under identical preprocessing, initialisation, and multi-start settings, with the true number of clusters supplied. This ensures consistency and comparability with the results reported earlier. Given the number of clusters is not known, we experimented with \(k \in [2, \lceil\sqrt{n}\rceil ]\)

\begin{figure}[hpt!]
    \centering
    \includegraphics[width=1.\linewidth]{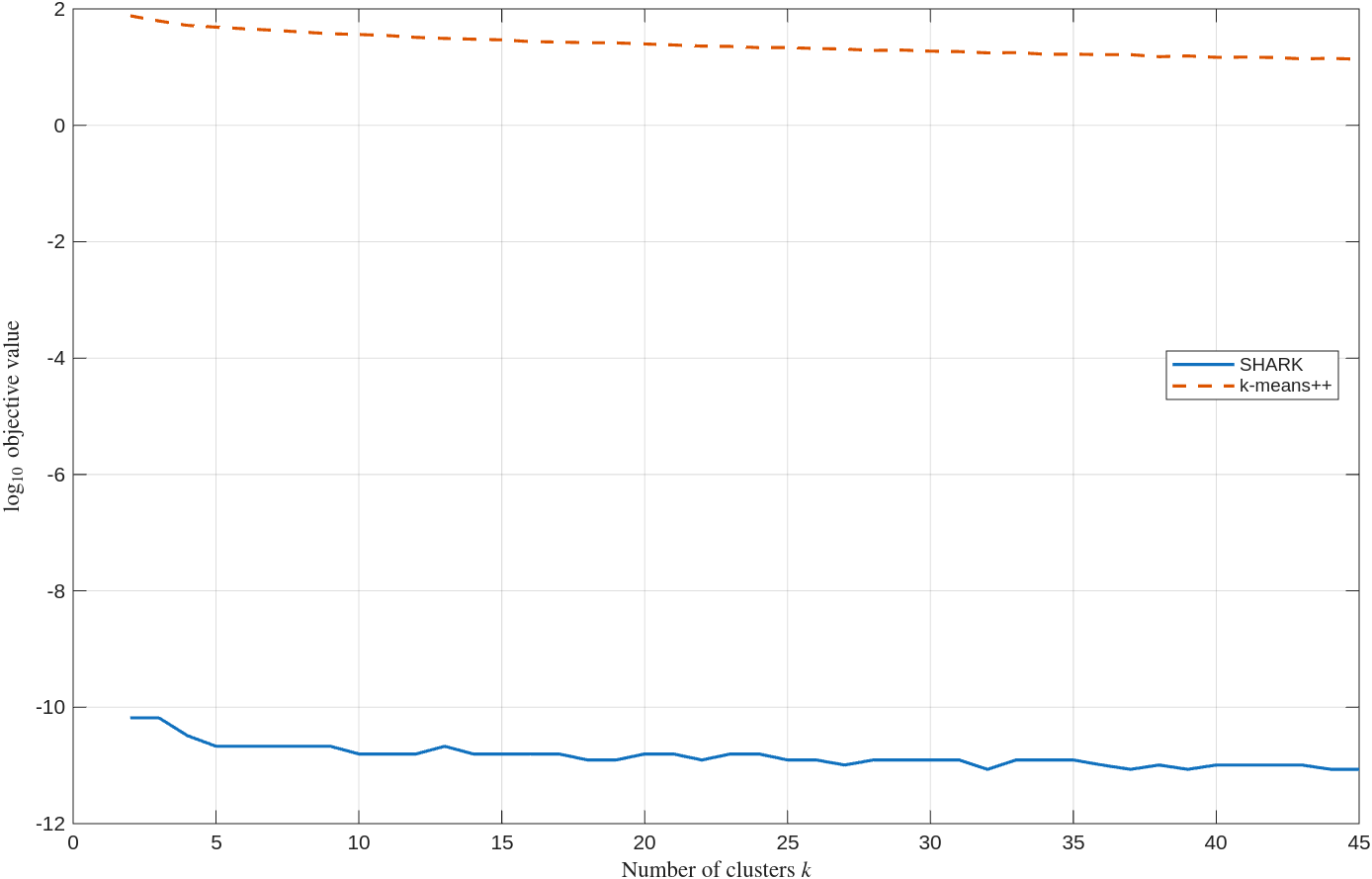}
    \caption{
        \textbf{Comparison of the (normalised) objective function values achieved by SHARK and $k$-means++ across different numbers of clusters, shown on a log$_{10}$ scale.}}
    \label{fig:malware_objective}
\end{figure}

Let us start by analysing objective functions. Figure~\ref{fig:malware_objective} compares the evolution of the objective function values attained by SHARK and $k$-means++ per value of $k$, reported on a logarithmic scale. Although both objective functions are comparable (see Observation~\ref{obs:comparable_kmeans}), SHARK consistently attains substantially lower objective values than $k$-means++ across all values of $k$. The large and persistent gap between the curves indicates that SHARK rapidly suppresses high-dispersion (less informative) features, leading to a markedly lower clustering cost even as the number of clusters increases.

Notably, both methods exhibit a gradual decrease in the objective as $k$ grows, reflecting the increased flexibility afforded by additional clusters. However, the relative ordering of the objectives remains unchanged, and the gap between SHARK and $k$-means++ does not diminish, suggesting that the advantage conferred by Shapley-based feature reweighting is stable across different model granularities.
This empirical behaviour is consistent with the theoretical results in Theorem~\ref{thm:SHARK_lowerbound}, which guarantee that SHARK achieves a strictly lower comparable objective whenever feature dispersions are heterogeneous.

\begin{figure}[hpt!]
    \centering
    \includegraphics[width=1.\linewidth]{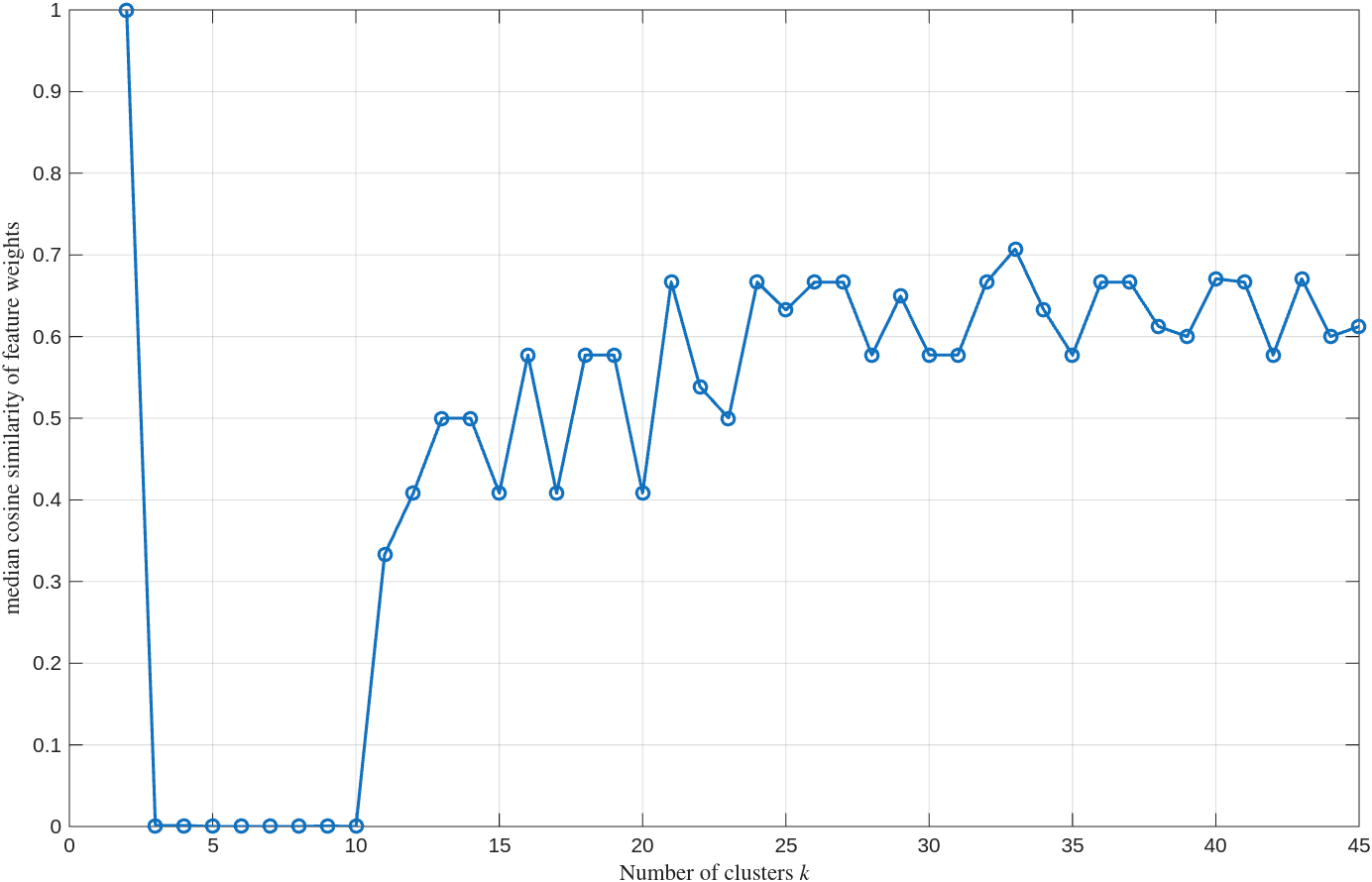}
    \caption{
        \textbf{Median cosine similarity of SHARK feature weight vectors across repeated runs for different numbers of clusters $k$, demonstrating the stability of the learned feature relevance structure in the malware data set.}}
    \label{fig:malware_weight}
\end{figure}

Let us now analyse the stability of the weights. Figure~\ref{fig:malware_weight} presents the median cosine similarity between feature-weight vectors obtained from the 25 runs of SHARK per value of $k$. For each value of \(k\), the cosine similarity was computed across all pairs of runs, and the median value is shown to summarise stability.

When \(k=2\), SHARK yields identical feature-weight vectors across most runs, resulting in a median cosine similarity of one. This reflects a highly stable weighting regime in which the algorithm consistently emphasises the same subset of behavioural and static features. In the context of malware analysis, this behaviour is consistent with prior findings reported in the original study introducing this data set~\citep{de2021identifying}, where coarse partitions often separate malware samples according to broad behavioural traits, such as dominant execution or communication patterns.

As the number of clusters increases, the clustering task becomes more granular and allows for multiple plausible partitions, which is reflected in a transient reduction in weight similarity for intermediate values of \(k\). Despite this increased flexibility, the median cosine similarity steadily increases and stabilises in the range of approximately \(0.6\)–\(0.7\) for larger values of \(k\). This indicates that SHARK consistently recovers similar feature-importance profiles across independent runs, even when modelling finer-grained malware groupings.

Overall, these results suggest that SHARK identifies stable and reproducible feature relevance structures that align with known characteristics of malware behaviour. The stability observed across a wide range of cluster numbers supports the use of SHARK as an exploratory tool for malware analysis, where the true number of families is typically unknown and robustness to initialisation is essential.

\section{Conclusion}

This paper introduced SHARK, a feature weighting clustering algorithm motivated by Shapley values. Like $k$-means, SHARK requires only the data set and the desired number of clusters, $k$, with no additional parameters. This is a significant practical advantage over most existing feature-weighted clustering algorithms, where parameter selection can be challenging and computationally expensive.

Theoretically, we established in Theorem~\ref{thm:shapley_kmeans} that the $k$-means objective can be decomposed exactly into the sum of per-feature Shapley values. This result provides an axiomatic foundation for quantifying feature relevance in an unsupervised setting and reveals that, for the $k$-means objective, Shapley values can be computed in closed form due to the additivity property (see Observation~\ref{obs:kmeans_additive}). This reduces the computational complexity from exponential, in the general Shapley framework, to polynomial time for our setting.

We showed how SHARK replaces the arithmetic mean of the feature dispersions, optimised by $k$-means, with their harmonic mean, leading to a guaranteed reduction in the comparable objective whenever features differ in dispersion (see Theorem~\ref{thm:SHARK_lowerbound}). This property underpins SHARK's ability to down-weight fewer features while preserving or amplifying the contribution of relevant ones, ensuring that its cost is never worse than that of the comparable $k$-means given the same centroids.

Empirically, SHARK consistently matches or outperforms benchmarked algorithms across diverse synthetic and real-world data sets, with particularly strong gains in high-dimensional and noisy conditions. These results confirm that SHARK’s theoretical properties (lack of additional parameters relative to $k$-means, fair and principled feature relevance quantification, and guaranteed improvement over comparable $k$-means) translate into robust practical performance. Together, these qualities position SHARK as a scalable, principled, and efficient solution for feature-weighted clustering.

In terms of future research, we intend to address a limitation shared with $k$-means: the need to pre-specify the number of clusters $k$. We find that investigating how SHARK can be combined with data-driven strategies for estimating or adapting $k$ (or how the choice of $k$ might be integrated into the feature-weighting framework itself) is a natural direction. Another promising direction is to further exploit SHARK’s feature weights for explainability, for instance by using them to construct structured, practitioner-oriented counterfactual explanations rather than simple ranked lists.

\clearpage
\cleardoublepage   

\begin{table*}[htb!]\small
\centering
\scalebox{0.88}{
\begin{tabular}{l r r r r r}
\toprule
 {Experiment} &  {$k$-means++} &  {FWSA} &  {L-W$k$-means} &  {KMTD} &  {SHARK} \\
\midrule
 {1000x10\_3k}        &  {00:00:01} &  {00:00:17} &  {00:00:38} &  {00:02:34} &  {00:00:06} \\
 {1000x10\_3k+5NF}    &  {00:00:07} &  {00:00:18} &  {00:02:44} &  {00:17:56} &  {00:00:07} \\
 {1000x10\_5k}        &  {00:00:05} &  {00:00:32} &  {00:02:06} &  {00:09:03} &  {00:00:11} \\
 {1000x10\_5k+5NF}    &  {00:00:27} &  {00:00:33} &  {00:05:18} &  {00:32:38} &  {00:00:16} \\
 {1000x10\_10k}       &  {00:00:10} &  {00:01:20} &  {00:06:21} &  {00:24:45} &  {00:00:25} \\
 {1000x10\_10k+5NF}   &  {00:00:10} &  {00:02:39} &  {00:13:27} &  {00:59:58} &  {00:00:51} \\
 {2000x20\_5k}        &  {00:00:06} &  {00:01:45} &  {00:07:03} &  {00:20:48} &  {00:00:17} \\
 {2000x20\_5k+10NF}   &  {00:00:21} &  {00:02:46} &  {00:53:08} &  {01:03:39} &  {00:00:21} \\
 {2000x20\_10k}       &  {00:00:16} &  {00:04:20} &  {00:27:03} &  {00:37:27} &  {00:00:36} \\
 {2000x20\_10k+10NF}  &  {00:00:26} &  {00:04:56} &  {01:26:24} &  {03:56:34} &  {00:00:41} \\
 {2000x20\_20k}       &  {00:00:20} &  {00:03:52} &  {00:28:59} &  {02:02:05} &  {00:01:23} \\
 {2000x20\_20k+10NF}  &  {00:00:20} &  {00:05:26} &  {01:11:44} &  {04:24:38} &  {00:01:41} \\
 {2000x30\_5k}        &  {00:00:02} &  {00:01:03} &  {00:05:57} &  {00:20:03} &  {00:00:29} \\
 {2000x30\_5k+15NF}   &  {00:00:05} &  {00:01:11} &  {00:14:24} &  {00:21:58} &  {00:00:31} \\
 {2000x30\_10k}       &  {00:00:08} &  {00:02:19} &  {00:11:48} &  {02:05:05} &  {00:00:41} \\
 {2000x30\_10k+15NF}  &  {00:00:23} &  {00:02:58} &  {01:14:20} &  {02:50:23} &  {00:01:03} \\
 {2000x30\_20k}       &  {00:00:18} &  {00:03:03} &  {00:30:33} &  {01:23:53} &  {00:01:18} \\
 {2000x30\_20k+15NF}  &  {00:00:28} &  {00:05:16} &  {01:23:59} &  {03:31:22} &  {00:02:13} \\
 {5000x50\_10k}       &  {00:00:13} &  {00:09:12} &  {00:57:53} &  {02:44:29} &  {00:07:02} \\
 {5000x50\_10k+25NF}  &  {00:00:16} &  {00:13:36} &  {02:00:19} &  {01:33:01} &  {00:09:33} \\
 {5000x50\_20k}       &  {00:00:25} &  {00:16:49} &  {02:41:58} &  {02:05:09} &  {00:12:27} \\
 {5000x50\_20k+25NF}  &  {00:01:13} &  {00:18:45} &  {10:15:25} &  {19:57:25} &  {00:18:11} \\
 {5000x50\_50k}       &  {00:01:06} &  {00:29:40} &  {08:16:42} &  {04:08:43} &  {00:29:24} \\
 {5000x50\_50k+25NF}  &  {00:01:31} &  {00:41:23} &  {01:35:52} &  {04:12:09} &  {00:39:19} \\
\bottomrule
\end{tabular}
} 
\caption{ {\textbf{Running time for experiments, baselined at single-threaded processing}. Times reflect the duration of each experiment, which included 25 iterations of each of 50 datasets, though some algorithms used additional retry logic when a failure of cluster recovery was intercepted. }}
\label{tab:running_time}
\end{table*}


\begin{table*}[htb!]\small
\centering
\scalebox{0.88}{
\begin{tabular}{l r r r r r}
\toprule
 {Experiment} &  {$k$-means++} &  {SHARK} &  {FWSA} &  {L-W$k$-means} &  {KMTD} \\
\midrule
 {1000x10\_3k}        &  {1}  &  {6}   &  {17}  &  {38}   &  {154}   \\
 {1000x10\_3k+5NF}    &  {7}  &  {7}   &  {18}  &  {164}  &  {1076}  \\
 {1000x10\_5k}        &  {5}  &  {11}  &  {32}  &  {126}  &  {543}   \\
 {1000x10\_5k+5NF}    &  {27} &  {16}   &  {33}  &  {318}  &  {1958} \\
 {1000x10\_10k}       &  {10} &  {25}   &  {80}  &  {381}  &  {1485} \\
 {1000x10\_10k+5NF}   &  {10} &  {51}  &  {159} &  {807}  &  {3598}  \\
 {2000x20\_5k}        &  {6}  &  {17}  &  {105} &  {423}  &  {1248}  \\
 {2000x20\_5k+10NF}   &  {21} &  {21}  &  {166} &  {3188} &  {3819}  \\
 {2000x20\_10k}       &  {16} &  {36}  &  {260} &  {1623} &  {2247}  \\
 {2000x20\_10k+10NF}  &  {26} &  {41}  &  {296} &  {5184} &  {14194} \\
 {2000x20\_20k}       &  {20} &  {83}  &  {232} &  {1739} &  {7325}  \\
 {2000x20\_20k+10NF}  &  {20} &  {101} &  {326} &  {4304} &  {15878} \\
 {2000x30\_5k}        &  {2}  &  {29}  &  {63}  &  {357}  &  {1203}  \\
 {2000x30\_5k+15NF}   &  {5}  &  {31}  &  {71}  &  {864}  &  {1318}  \\
 {2000x30\_10k}       &  {8}  &  {41}  &  {139} &  {708}  &  {7505}  \\
 {2000x30\_10k+15NF}  &  {23} &  {63}  &  {178}  &  {4460} &  {10223} \\
 {2000x30\_20k}       &  {18} &  {78}  &  {183} &  {1833} &  {5033}  \\
 {2000x30\_20k+15NF}  &  {28} &  {133} &  {316} &  {5039} &  {12682} \\
 {5000x50\_10k}       &  {13} &  {422} &  {552} &  {3473} &  {9869}  \\
 {5000x50\_10k+25NF}  &  {16} &  {573} &  {816}  &  {7219}  &  {5581} \\
 {5000x50\_20k}       &  {25} &  {747} &  {1009} &  {9718} &  {7509}  \\
 {5000x50\_20k+25NF}  &  {73} &  {1091} &  {1125} &  {36925} &  {71645} \\
 {5000x50\_50k}       &  {66} &  {1764} &  {1780} &  {29802} &  {14923} \\
 {5000x50\_50k+25NF}  &  {91} &  {2359} &  {2483} &  {5752}  &  {15129} \\
\bottomrule
\end{tabular}
}
\caption{ {\textbf{Normalised running times for all experiments}. 
The baseline was set such that $k$-means++ on 1000x10\_3k = 1). \\ }} 
\label{tab:running_time_normalised}
\end{table*}

\clearpage

\ifarxiv
    \bibliographystyle{apalike}  
    \bibliography{references}   

@article{ran2023comprehensive,
  title={Comprehensive survey on hierarchical clustering algorithms and the recent developments},
  author={Ran, Xingcheng and Xi, Yue and Lu, Yonggang and Wang, Xiangwen and Lu, Zhenyu},
  journal={Artificial Intelligence Review},
  volume={56},
  number={8},
  pages={8219--8264},
  year={2023},
  publisher={Springer}
}

@inproceedings{de2012empirical,
  title={An empirical evaluation of different initializations on the number of k-means iterations},
  author={De Amorim, Renato Cordeiro},
  booktitle={Mexican international conference on artificial intelligence},
  pages={15--26},
  year={2012},
  organization={Springer}
}

@article{mahdi2021scalable,
  title={Scalable clustering algorithms for big data: A review},
  author={Mahdi, Mahmoud A and Hosny, Khalid M and Elhenawy, Ibrahim},
  journal={IEEE Access},
  volume={9},
  pages={80015--80027},
  year={2021},
  publisher={IEEE}
}

@software{MATLAB,
year = {2022},
author = {{The MathWorks Inc.}},
title = {MATLAB version: 9.13.0 (R2022b)},
publisher = {The MathWorks Inc.},
address = {Natick, Massachusetts, United States},
url = {https://www.mathworks.com}
}

@article{marcilio2021explaining,
  title={Explaining dimensionality reduction results using Shapley values},
  author={Marcilio-Jr, Wilson E and Eler, Danilo M},
  journal={Expert Systems with Applications},
  volume={178},
  pages={115020},
  year={2021},
  publisher={Elsevier}
}

@article{sharma2008some,
  title        = {Some more inequalities for arithmetic mean, harmonic mean and variance},
  author       = {Sharma, Rajesh},
  journal      = {Journal of Mathematical Inequalities},
  volume       = {2},
  number       = {2},
  pages        = {239--243},
  year         = {2008}
}

@article{al2024genetic,
  title={Genetic programming for feature selection based on feature removal impact in high-dimensional symbolic regression},
  author={Al-Helali, Baligh and Chen, Qi and Xue, Bing and Zhang, Mengjie},
  journal={IEEE Transactions on Emerging Topics in Computational Intelligence},
  year={2024},
  publisher={IEEE}
}

@article{yin2022adaptive,
  title={Adaptive feature selection with shapley and hypothetical testing: Case study of EEG feature engineering},
  author={Yin, Dingze and Chen, Dan and Tang, Yunbo and Dong, Heyou and Li, Xiaoli},
  journal={Information Sciences},
  volume={586},
  pages={374--390},
  year={2022},
  publisher={Elsevier}
}

@manual{RLanguage,
     title = {R: A Language and Environment for Statistical Computing},
     author = {{R Core Team}},
     organization = {R Foundation for Statistical Computing},
     address = {Vienna, Austria},
     year = {2021},
     url = {https://www.R-project.org/},
   }

@article{de2021identifying,
  title={Identifying meaningful clusters in malware data},
  author={de Amorim, Renato Cordeiro and Ruiz, Carlos David Lopez},
  journal={Expert Systems with Applications},
  volume={177},
  pages={114971},
  year={2021},
  publisher={Elsevier}
}

@article{scikit-learn,
  title={Scikit-learn: Machine Learning in {P}ython},
  author={Pedregosa, F. and Varoquaux, G. and Gramfort, A. and Michel, V.
          and Thirion, B. and Grisel, O. and Blondel, M. and Prettenhofer, P.
          and Weiss, R. and Dubourg, V. and Vanderplas, J. and Passos, A. and
          Cournapeau, D. and Brucher, M. and Perrot, M. and Duchesnay, E.},
  journal={Journal of Machine Learning Research},
  volume={12},
  pages={2825--2830},
  year={2011}
}

@article{kumar2024high,
  title={High-density cluster core-based k-means clustering with an unknown number of clusters},
  author={Kumar, Abhimanyu and Kumar, Abhishek and Mallipeddi, Rammohan and Lee, Dong-Gyu},
  journal={Applied Soft Computing},
  volume={155},
  pages={111419},
  year={2024},
  publisher={Elsevier}
}

@article{zhang2024speeding,
  title={Speeding up k-means clustering in high dimensions by pruning unnecessary distance computations},
  author={Zhang, Haowen and Li, Jing and Zhang, Junru and Dong, Yabo},
  journal={Knowledge-Based Systems},
  volume={284},
  pages={111262},
  year={2024},
  publisher={Elsevier}
}

@article{zhang2025structured,
  title={Structured multi-view k-means clustering},
  author={Zhang, Zitong and Chen, Xiaojun and Wang, Chen and Wang, Ruili and Song, Wei and Nie, Feiping},
  journal={Pattern Recognition},
  volume={160},
  pages={111113},
  year={2025},
  publisher={Elsevier}
}

@article{sinaga2020unsupervised,
  title={Unsupervised K-means clustering algorithm},
  author={Sinaga, Kristina P and Yang, Miin-Shen},
  journal={IEEE access},
  volume={8},
  pages={80716--80727},
  year={2020},
  publisher={IEEE}
}

@article{shetty2021hierarchical,
  title={Hierarchical clustering: a survey},
  author={Shetty, Pranav and Singh, Suraj},
  journal={International Journal of Applied Research},
  volume={7},
  number={4},
  pages={178--181},
  year={2021}
}

@article{oyewole2023data,
  title={Data clustering: application and trends},
  author={Oyewole, Gbeminiyi John and Thopil, George Alex},
  journal={Artificial intelligence review},
  volume={56},
  number={7},
  pages={6439--6475},
  year={2023},
  publisher={Springer}
}

@article{bushra2021comparative,
  title={Comparative analysis review of pioneering DBSCAN and successive density-based clustering algorithms},
  author={Bushra, Adil Abdu and Yi, Gangman},
  journal={IEEE Access},
  volume={9},
  pages={87918--87935},
  year={2021},
  publisher={IEEE}
}

@article{bhattacharjee2021survey,
  title={A survey of density based clustering algorithms},
  author={Bhattacharjee, Panthadeep and Mitra, Pinaki},
  journal={Frontiers of Computer Science},
  volume={15},
  pages={1--27},
  year={2021},
  publisher={Springer}
}

@article{chakraborty2022,
  author = {Chakraborty, Saptarshi and Das, Swagatam},
  title = {Detecting Meaningful Clusters From High-Dimensional Data: A Strongly Consistent Sparse Center-Based Clustering Approach},
  journal = {IEEE Transactions on Knowledge and Data Engineering},
  volume = {44},
  number = {6},
  pages = {2894--2908},
  year = {2022},
  url = {https://ieeexplore.ieee.org/document/9309172}
}

@inproceedings{macqueen1967,
  author = {MacQueen, James},
  title = {Some methods for classification and analysis of multivariate observations},
  booktitle = {Proceedings of the Fifth Berkeley Symposium on Mathematical Statistics and Probability, Volume 1: Statistics},
  volume = {5},
  pages = {281--298},
  year = {1967},
  organization = {University of California Press}
}

@article{tsai2008developing,
  author = {Tsai, Chieh-Yuan and Chiu, Chuang-Cheng},
  title = {Developing a feature weight self-adjustment mechanism for a k-means clustering algorithm},
  journal = {Computational Statistics \& Data Analysis},
  volume = {52},
  number = {10},
  pages = {4658--4672},
  year = {2008},
  publisher = {Elsevier}
}

@article{ikotun2023k,
  title={K-means clustering algorithms: A comprehensive review, variants analysis, and advances in the era of big data},
  author={Ikotun, Abiodun M and Ezugwu, Absalom E and Abualigah, Laith and Abuhaija, Belal and Heming, Jia},
  journal={Information Sciences},
  volume={622},
  pages={178--210},
  year={2023},
  publisher={Elsevier}
}

@article{deng2016survey,
  title={A survey on soft subspace clustering},
  author={Deng, Zhaohong and Choi, Kup-Sze and Jiang, Yizhang and Wang, Jun and Wang, Shitong},
  journal={Information sciences},
  volume={348},
  pages={84--106},
  year={2016},
  publisher={Elsevier}
}

@article{shapley1953value,
  title={A value for n-person games},
  author={Shapley, Lloyd S and others},
  year={1953},
  publisher={Princeton University Press Princeton}
}

@article{hubert1985comparing,
  title={Comparing partitions},
  author={Hubert, Lawrence and Arabie, Phipps},
  journal={Journal of Classification},
  volume={2},
  number={1},
  pages={193--218},
  year={1985},
  publisher={Springer}
}

@article{harris2022extensive,
  title={An extensive empirical comparison of k-means initialization algorithms},
  author={Harris, Simon and De Amorim, Renato Cordeiro},
  journal={IEEE Access},
  volume={10},
  pages={58752--58768},
  year={2022},
  publisher={IEEE}
}

@article{hancer2020survey,
  title={A survey on feature selection approaches for clustering},
  author={Hancer, Emrah and Xue, Bing and Zhang, Mengjie},
  journal={Artificial Intelligence Review},
  volume={53},
  number={6},
  pages={4519--4545},
  year={2020},
  publisher={Springer}
}

@article{xue2024remote,
  title={Remote Parkinson's disease severity prediction based on causal game feature selection},
  author={Xue, Zaifa and Lu, Huibin and Zhang, Tao and Guo, Xiaonan and Gao, Le},
  journal={Expert Systems with Applications},
  volume={241},
  pages={122690},
  year={2024},
  publisher={Elsevier}
}

@article{li2024unsupervised,
  title={Unsupervised feature selection using chronological fitting with Shapley Additive explanation (SHAP) for industrial time-series anomaly detection},
  author={Li, Qixuan and Ji, Yangjian and Zhu, Mingrui and Zhu, Xiaoyang and Sun, Linjin},
  journal={Applied Soft Computing},
  volume={155},
  pages={111426},
  year={2024},
  publisher={Elsevier}
}

@inproceedings{vassilvitskii2006k,
  title={k-means++: The advantages of careful seeding},
  author={Vassilvitskii, Sergei and Arthur, David},
  booktitle={Proceedings of the eighteenth annual ACM-SIAM symposium on Discrete algorithms},
  pages={1027--1035},
  year={2006}
}

@article{huang2005ewkm,
  author  = {Huang, Jian-Zhong and Ng, Michael K. and Rong, Hongqiang and Li, Zhirong},
  title   = {Automated Variable Weighting in k-Means Type Clustering},
  journal = {IEEE Transactions on Pattern Analysis and Machine Intelligence},
  volume  = {27},
  number  = {5},
  pages   = {657--668},
  year    = {2005}
}

@article{Rousseeuw1987,
  author    = {Rousseeuw, Peter J.},
  title     = {Silhouettes: A graphical aid to the interpretation and validation of cluster analysis},
  journal   = {Journal of Computational and Applied Mathematics},
  volume    = {20},
  pages     = {53--65},
  year      = {1987},
  doi       = {10.1016/0377-0427(87)90125-7}
}

@article{xiao2025robust,
  title={Robust k-means-type Clustering for Noisy Data},
  author={Xiao, Han and Li, Peng and Wei, Yubin and Yu, Ning and Fang, Yong},
  journal={IEEE Transactions on Pattern Analysis and Machine Intelligence},
  year={2025}
}

@misc{uci2019,
  author = {Dua, Dheeru and Graff, Casey},
  title = {UCI Machine Learning Repository},
  year = {2019},
  institution = {University of California, Irvine},
  howpublished = {\url{https://archive.ics.uci.edu/ml}},
}

@article{peel2000robust,
  author = {Peel, D. and McLachlan, G. J.},
  title = {Robust mixture modelling using t distribution},
  journal = {Statistics and Computing},
  volume = {10},
  number = {4},
  pages = {339--348},
  year = {2000}
}
\fi

\end{document}